\documentclass[sigconf,nonacm]{acmart}

\AtBeginDocument{%
  \providecommand\BibTeX{{%
    \normalfont B\kern-0.5em{\scshape i\kern-0.25em b}\kern-0.8em\TeX}}}

\acmConference[Preprint]{Preprint}
\usepackage{comment}
\begin{document}

\title{KSAT: \underline{K}nowledge-infused \underline{S}elf \underline{A}ttention \underline{T}ransformer - Integrating Multiple Domain-Specific Contexts}

\author{Kaushik Roy}
\email{kaushikr@email.sc.edu}
\affiliation{%
  \institution{Artificial Intelligence Institute \\ University of South Carolina}
  \country{USA}
}
\author{Yuxin Zi}
\email{yzi@email.sc.edu}
\affiliation{%
  \institution{Artificial Intelligence Institute \\ University of South Carolina}
  \country{USA}
}
\author{Vignesh Narayanan}
\email{vignar@sc.edu}
\affiliation{%
  \institution{Artificial Intelligence Institute \\ University of South Carolina}
  \country{USA}
}
\author{Manas Gaur}
\email{manas@umbc.edu}
\affiliation{%
  \institution{KAI\textsuperscript{2}, University of Maryland \\ Baltimore County}
  \country{USA}
}
\author{Amit Sheth}
\email{amit@sc.edu}
\affiliation{%
  \institution{Artificial Intelligence Institute \\ University of South Carolina}
  \country{USA}
}

\begin{abstract}
Domain-specific language understanding requires integrating multiple pieces of relevant contextual information. For example,  we see both suicide and depression-related behavior (multiple contexts) in the text ``\textit{I have a gun and feel pretty bad about my life, and it wouldn't be the worst thing if I didn't wake up tomorrow}''. Domain specificity in self-attention architectures is handled by fine-tuning on excerpts from relevant domain specific resources (datasets and external knowledge - medical textbook chapters on mental health diagnosis related to suicide and depression). We propose a modified self-attention architecture \textit{Knowledge-infused Self Attention Transformer} (KSAT) that achieves the integration of multiple domain-specific contexts through the use of external knowledge sources. KSAT introduces knowledge-guided biases in dedicated self-attention layers for each knowledge source to accomplish this. In addition, KSAT provides mechanics for controlling the trade-off between learning from data and learning from knowledge. Our quantitative and qualitative evaluations show that (1) the KSAT architecture provides novel human-understandable ways to precisely measure and visualize the contributions of the infused domain contexts, and (2) KSAT performs competitively with other knowledge-infused baselines and significantly outperforms baselines that use fine-tuning for domain-specific tasks. 
\end{abstract}





\maketitle
\section{Motivation}\label{sec:intro}
Solving domain-specific tasks such as mental health diagnosis (MHD), and triaging, requires integrating relevant contextual information from data and knowledge sources. Self-Attention based Language Models (SAMs) capture an aggregated broader context from domain-agnostic, voluminous training corpora \cite{devlin2018bert}. Fine-tuning SAMs on domain-specific corpora achieves domain-specific context capture \cite{sun2019fine,rasmy2021med}. However, SAM architectures are black-box in nature \cite{gaur2021semantics}. Consequently, fine-tuned SAM architectures do not lend themselves to the robust evaluation of the open research aims: (\textbf{R1}) 
Relevant domain-specific context coverage, and (\textbf{R2}) The influence of knowledge context traded-off against the data context in downstream tasks \cite{bellegarda2004statistical,gururangan2020don}. We propose a modified self-attention architecture \textit{Knowledge-infused Self Attention Transformer} (KSAT) to address these aims. KSAT performs well on select domain-specific tasks (see \ref{subsec:resources}) while lending itself to a robust human-understandable evaluation of \textbf{R1} and \textbf{R2}. Thus KSAT provides a substantial step towards fostering AI-user trust, and satisfaction \cite{sheth2021knowledge,sheth2022process}.
\section{Background}\label{sec:background}
\subsection{Related Work}\label{subsec:rel_work}
Prior approaches that are relevant to \textbf{R1} and \textbf{R2} and incorporate multiple knowledge contexts can be broadly categorized based on the knowledge-infusion technique as (\textbf{1}) knowledge modulated SAMs and (\textbf{2}) knowledge infused input embedding-based SAMs \cite{peters2019knowledge,wang2020k}. The former uses knowledge to guide the self-attention mechanism in SAMs, and the latter embeds the knowledge into a vector space before passing the inputs into SAMs. Here, we briefly summarize their contributions towards \textbf{R1} and \textbf{R2}. Both Category (\textbf{1}), and Category (\textbf{2}) methods' domain coverage is evaluated through performance on domain-specific task descriptions (\textbf{R1}). These methods' ablations highlight contributions of knowledge context (\textbf{R2}). However, inspecting the numerical outputs from the model components (projection matrices and vectors) does not easily lend themselves to human-understandable scrutiny. Explainable AI techniques (post-processing of the numerical outputs that transform them into human-understandable information) are required to confirm the author(s) perspectives \cite{xu2019explainable}. Post-processing-based explanations are local approximations of the SAM reasoning for particular inputs and therefore do not present the global picture, casting doubts on the SAM evaluation validity. KSAT presents a SAM architecture whose numerical outputs lend themselves to robust human-understandable evaluations of \textbf{R1} and \textbf{R2}.

\subsection{Task Description, Data, and External Knowledge Sources}\label{subsec:resources}
\begin{figure}[!h]
    \centering
    \includegraphics[width=\linewidth,keepaspectratio]{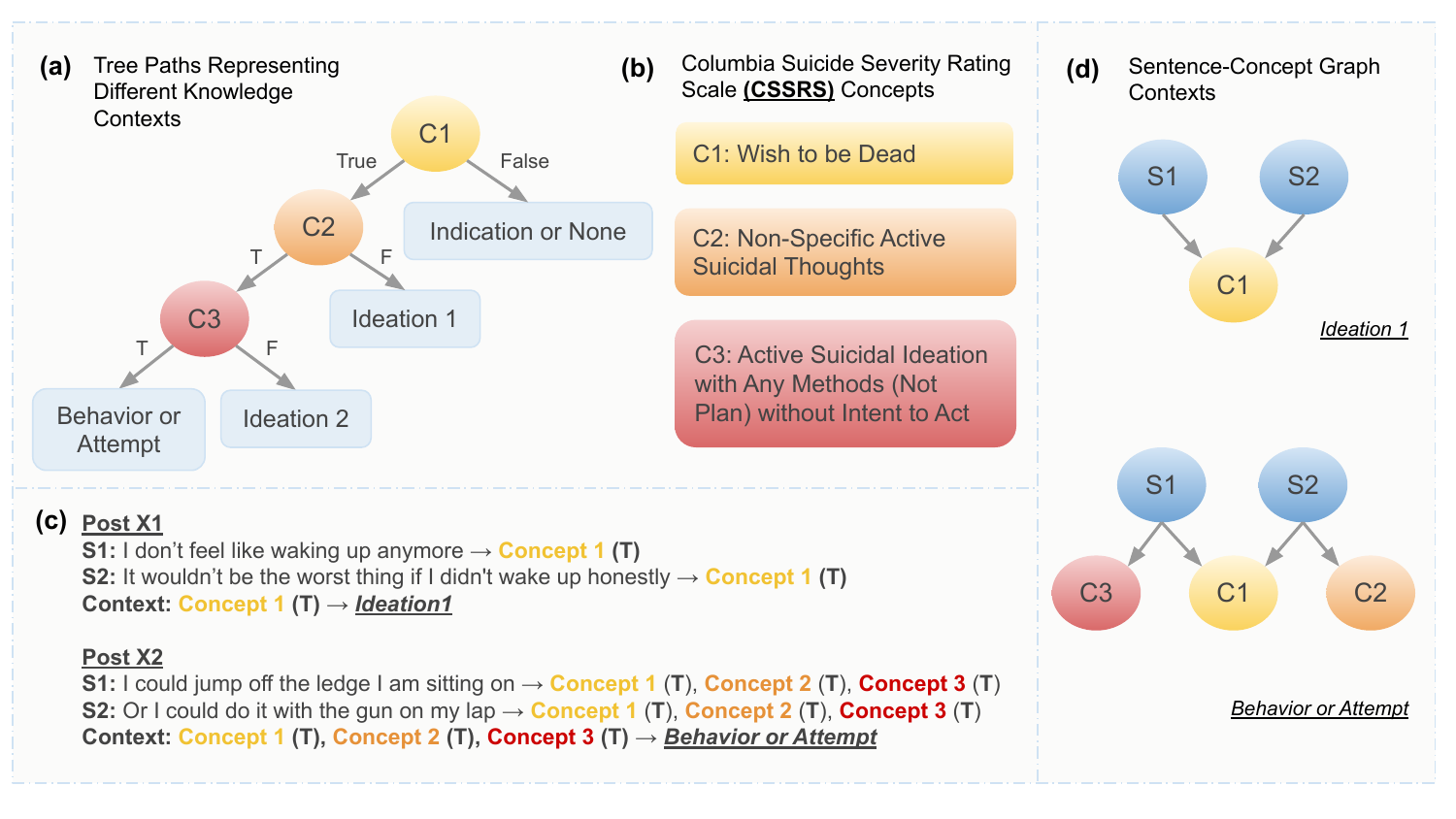}
    \caption{\footnotesize The knowledge source we utilize for our experiments is the \textit{Columbia Suicide Severity Rating Scale} (CSSRS) and the data are posts from Reddit subreddits related to suicide. (\textbf{a}) Shows the knowledge contexts (tree-path represents a context) present in the CSSRS leading to varying suicidal patterns. (\textbf{b}) Shows the different concepts listed in the CSSRS. (\textbf{c}) Shows the posts and sentences from the data annotated with relevant knowledge concepts and contexts from the CSSRS. (\textbf{d}) Shows how sentences from the input posts (\textbf{Post X1} and \textbf{Post X2}) and concepts from the  CSSRS are graphically connected in different contexts (In this case, the \underline{\textit{Ideation1}} and the \underline{\textit{Behavior or Attempt}} contexts).}
    \label{fig:cssrs}
\end{figure}

Although the KSAT architecture broadly applies to any domain-specific task, we choose the specific task of Mental Health Diagnostic Assistance for Suicidal Tendencies by Gaur et al. \cite{gaur2019knowledge}. We denote this dataset as MHDA. The data contains high-quality expert annotations on Reddit posts from suicide-related subreddits. The annotation method ensures minimal noise from measurement artifacts and high agreement among the expert annotators. We use the clinically established diagnostic process information contained in the \textit{Columbia Suicide Severity Rating Scale} (CSSRS) for knowledge contexts. Figure \ref{fig:cssrs} (a, b) illustrate the various contexts (each tree path represents a context) under which suicidal patterns can arise. The task is to predict the suicidal patterns, namely - \textit{indication}, \textit{ideation1}, \textit{ideation2}, and \textit{behavior or attempt}. Figure \ref{fig:cssrs} (c) shows examples from the MHDA dataset, augmented with knowledge context annotations. We denote the augmented dataset as k-MHDA. k-MHDA contains knowledge context annotations at the post and sentence level (see Figure \ref{fig:cssrs} (c)). We defer construction details of k-MHDA from the CSSRS knowledge and the MHDA data to the appendix Section \ref{subsec:kmhda} 
as it is not the main focus of the paper\footnote{We will release the k-MHDA dataset, along with code to construct it along with the KSAT code for reproducibility of results.}.
\section{KSAT - Proposed Architecture}\label{sec:ksat_arch}
For an input $x$ in the vanilla SAM layer $l$, a $CLS$ token is introduced that encodes the information in the data for downstream tasks. For example, for classification, $Z_{CLS}$, the representation of the $CLS$ token from the final layer $L$ is passed through a softmax layer which outputs class probabilities. In the KSAT layer, we introduce a $Z_{KCLS}$ token that encodes the knowledge for that layer. The $Z_{KCLS}$ values are determined by the graph context between the input $x$ sentences and the concepts in the knowledge context annotations for input $x$. As an example, consider the inputs  $x$ to be like the posts \textbf{Post X1} and \textbf{Post X2} shown in Figure \ref{fig:cssrs}
(c), then the graphs in Figure \ref{fig:cssrs} (d), illustrate the graph contexts that show sentence-concept connections. One would expect that the $Z_{KCLS}$ encoding for the sentences \textbf{S1} and \textbf{S2} in both posts would be similar as they have similar graph contexts. The $\mathbf{KG_{bias}}$ term  ensures that the values in $Z_{KCLS}$ captures this behavior (see Figure \ref{fig:ksat_arch} (b)).

\subsection{The KSAT layer}\label{sub:ksat_layer}
In contrast with the vanilla SAM layer, every KSAT layer has the following key differences:
\subsubsection{\textbf{Layer Parameters and Outputs}} \label{subsubsec:layer_output} 
\begin{figure}[!ht]
    \centering
    \includegraphics[width=\linewidth]{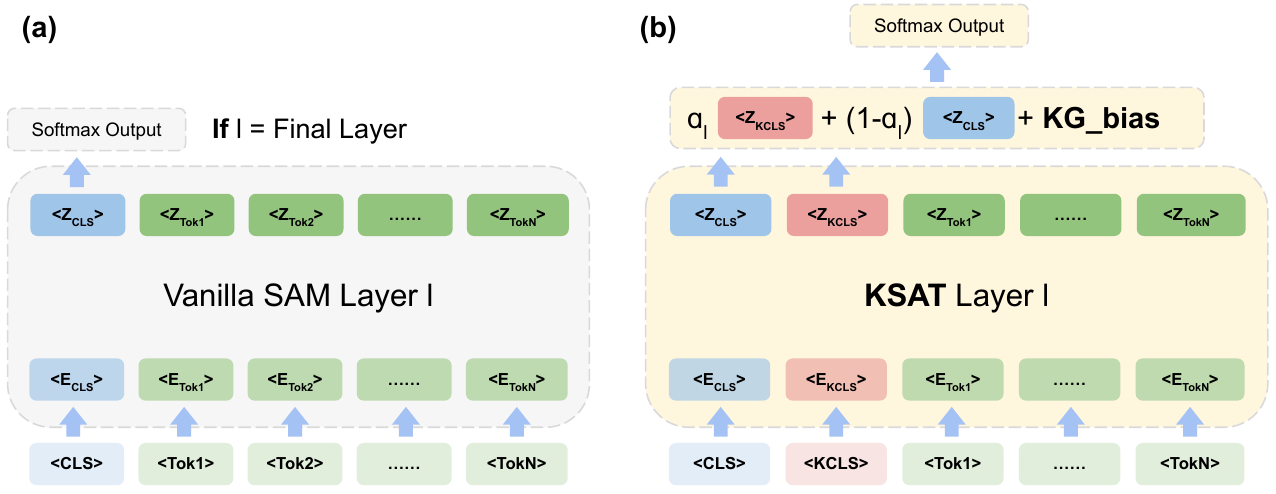}
    \caption{\footnotesize The comparative architecture of KSAT with existing vanilla SAMs. (a) Shows the vanilla SAM layer where the $Z_{CLS}$ token from the final layer is passed as input to the softmax function for prediction. (b) Shows the introduction of the additional $Z_{KCLS}$ token in a KSAT layer. The convex combination added with the knowledge bias as input to the softmax function is used to balance the use of data and knowledge contexts selectively.}
    \label{fig:ksat_arch}
\end{figure}

Let $Y$ denote the set of suicidal outcomes pertaining to each context: \{\textit{Indication or None}, \textit{Ideation1}, \textit{Ideation2}, \textit{Behavior or Attempt}\} (see Figure \ref{fig:cssrs} (a)). Let $x$ denote an input post (see Figure \ref{fig:cssrs} (c) for example posts). For each input $x$, every KSAT layer $l$ outputs a vector of probabilities for every outcome $y \in Y$ given by \textbf{Equation} \ref{eqn:layer_prob}:
\begin{equation}
    P_{l}^{ksat}(y|x) = \sigma(W^T(\alpha_{l} Z_{KCLS}(x) + (1-\alpha_{l})Z_{CLS}(x)) + \mathbf{KG_{bias}}(x))
    \label{eqn:layer_prob}
\end{equation}
$W$, $Z_{CLS}(x)$, and $Z_{KCLS}(x)$ are of dimension size $384 \times 1$ as we use the sentence-transformer published by Reimers et al. fine-tuned on the MHDA corpus, for embedding inputs \cite{reimers-2019-sentence-bert} . Recall that the term $\mathbf{KG_{bias}}$ is used to ensure that $Z_{CLS}(x)$ encodes the knowledge context represented as a graph (details in \ref{subsubsec:kg_bias}). The dimension size  of $\mathbf{KG_{bias}}$ is $1 \times 1$. The scalar term $\alpha_{l}$ is used 
as a trade-off factor modulating the contributions between data and knowledge contexts using a convex combination (see Equation \ref{eqn:layer_prob}). We contrast KSAT layer outputs against the vanilla SAM layer, where a similar probability vector is output only in the final layer and not at every layer (see Figure \ref{fig:ksat_arch}). The parameters within the vanilla SAM layer (the query, key, and value projection matrices) used to compute the self-attention matrix are retained in the KSAT layer \cite{devlin2018bert}. Unlike in the vanilla SAM layer, where the parameters may or may not be shared across layers, the parameters are not shared across different KSAT layers as different layers encode different knowledge contexts. 
\subsubsection{Encoding Graph Contexts in Layers}\label{subsubsec:kg_bias}
Even though every KSAT layer $l$ outputs a vector of probabilities for every outcome $y \in $
\{\textit{Indication or None}, \textit{Ideation1}, \textit{Ideation2}, \textit{Behavior or Attempt}\}, representing each context, it encodes only a single context in $Z_{KCLS}$. We show how to compute the $\mathbf{KG_{bias}}$ term by referencing the posts \textbf{Post X1} and \textbf{Post X2} and the graph contexts  \underline{\textit{Ideation1}} and \underline{\textit{Behavior or Attempt}} from Figure \ref{fig:cssrs} (c,d). 

\paragraph{\textbf{Computing Sentence-Concept Connection Vectors}:} We first compute a sentence-concept connection vector for each sentence in the posts. For \textbf{Post X1}, both sentence \textbf{S1} and \textbf{S2} are connected to the concept C1 in the concept set \{C1,C2,C3\} (see Figure \ref{fig:cssrs} (b)). Therefore both their sentence-concept connection vectors are computed as: [1, 0, 0]. Similarly for \textbf{Post X2}, the sentence-concept connection vectors for both \textbf{S1} and \textbf{S2} are computed as: [1, 1, 1]. 

\paragraph{\textbf{Computing Graph Context Distances}:}Recall that KSAT layers take as input a single post ($x$ in Equation \ref{eqn:layer_prob}). Denoting the sentence-concept vectors for a sentence \textbf{S1} as $C_{S1}$, to compute $\mathbf{KG_{bias}}(x)$, we first need to compute graph context distances for the posts \textbf{Post X1} and \textbf{Post X2}: $d(C_{S1},C_{S2})$. If the posts had more than two sentences the graph context distances would include all pairs (Eg: $d(C_{S1},C_{S2})$, $d(C_{S2},C_{S3})$, $d(C_{S1},C_{S3})$ for three sentences). The term $d(C_{Si},C_{Sj})$ for a pair of sentences $(S_i,S_j)$ in post $x$ captures the graph context-based distance between the sentences. Intuitively, sentences that have equivalent graph contexts should have $d(S_i,S_j) = 0$. We use hamming distance in our experiments.

The $\mathbf{KG_{bias}}$ term for an input $x$ is thus given by Equation \ref{eqn:kg_bias} as:
\begin{equation}
    \mathbf{KG_{bias}}(x) = -\sum_{(S_i,S_j) \in x}\frac{(Z_{KCLS}(x)[S_i]-Z_{KCLS}(x)[S_j])^2}{d(S_i,S_j) + \epsilon}
    \label{eqn:kg_bias}
\end{equation}
The formulation for the $\mathbf{KG_{bias}}(x)$ term in Equation \ref{eqn:kg_bias} encourages the $Z_{KCLS}$ representations of sentences that are in the same graph context to be similar. The $\varepsilon$ term is to prevent dividing by zero errors.
\subsection{Aggregating KSAT Layer Outputs}
Combining KSAT layer probabilities given by Equation \ref{eqn:layer_prob} is application domain dependent. For the suicidal outcomes in the set: \{\textit{Indication or None}, \textit{Ideation1}, \textit{Ideation2}, \textit{Behavior or Attempt}\}, it is reasonable to expect that suicidal ideation (both \textit{Ideation1} and \textit{Ideation2}) precedes the act of attempting suicide (\textit{Behavior or Attempt}). In other application domains, the contexts could be independent of each other. 

In our experiments, we stack four KSAT layers corresponding to the outcomes \textit{Indication or None}, \textit{Ideation1}, \textit{Ideation2}, \textit{Behavior or Attempt}, in that order (the order is derived from the tree structure in Figure \ref{fig:cssrs} (a)). Typically, for dependent probability outcomes, $X$, and $Y$, where $X$ precedes $Y$, the probability $P(X,Y)$ would be modeled as $P(Y \mid X)P(X)$. However, since we stack KSAT layers in a particular order, we use a product approximation ($P(X,Y) = P(X)P(Y)$) as information is propagated upwards through the KSAT layers. Thus, the final layer probabilities from the KSAT layers is computed using Equation \ref{eqn:ksat_prob} as:
\begin{equation}
    P_{final}^{ksat}(y|x) = \prod_{l} P_l^{ksat}(y|x),
    \label{eqn:ksat_prob}
\end{equation} where $P_l^{ksat}(y|x)$ is given by Equation \ref{eqn:layer_prob} 

\section{KSAT - Results and Analysis}
Recall the research aims that KSAT addresses when integrating multiple contexts - (\textbf{R1}) Relevant domain-specific context coverage, and (\textbf{R2}) The influence of knowledge context traded-off against the data context in downstream tasks (see \textbf{Section} \ref{sec:intro}).
\subsection{KSAT - Quantitative Results (Addresses R1)}\label{subsec:quant}
Table \ref{tab:results} shows the accuracy / AUC-ROC scores (rounded-of) for KSAT vs two best-performing fine-tuned transformer models, and knowledge infused baseline models \textbf{K-type(i)} and \textbf{K-type(ii)} (see section \ref{subsec:rel_work}).  Due to space concerns we describe details of the models \textbf{K-type(i)} and \textbf{K-type(ii)} in the Appendix Section \ref{sec:baselines}. \textbf{KSAT} outperforms the fine-tuned transformer models and performs comparably with models \textbf{K-type(i)} and \textbf{K-type(ii)} on the k-MHDA dataset (see section \ref{subsec:kmhda}).
\begin{table}[!h]
    \centering
    \begin{tabular}{||c|c|c|c|c|c||}
        \hline
         \textbf{Dataset} & \textbf{KSAT} & \textbf{K1} & \textbf{K2} & \textbf{XLNET} & \textbf{RoBERTa}\\
         \hline \hline
         k-MHDA (Acc) & 83\% & 84\%& 84\%& 68\% & 68\%\\
         \hline
         k-MHDA (AUC) & 78 & 71 & 72 & 57 & 63\\
         \hline
    \end{tabular}
    \caption{\footnotesize Shows the accuracy / AUC-ROC scores (rounded-of) for KSAT vs fine-tuned transformer models, and knowledge infused baseline models \textbf{K-type(i)} (K1) and \textbf{K-type(ii)} (K2). \textbf{KSAT} outperforms all fine-tuned transformer models and performs comparably with models \textbf{K-type(i)} and \textbf{K-type(ii)} on the k-MHDA dataset.}
    \label{tab:results}
\end{table}
\subsection{KSAT-Qualitative Results (Addresses R2)}\label{subsec:qual}
\begin{figure}[!h]
    \centering
    \includegraphics[trim = 3cm 5cm 4.0cm 3cm, width=\linewidth, clip]{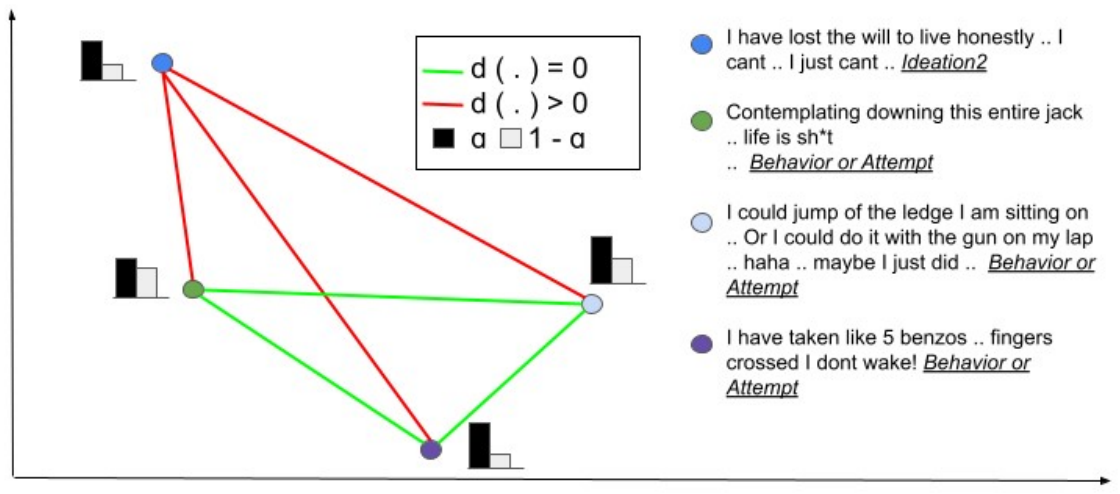}
    \caption{\footnotesize Shows that the first two \textit{Behavior or Attempt} posts in the list of posts are far off based on the $Z_{CLS}$ representations (\textcolor{red}{false}), whereas KSAT representations depict them as closer i.e d(.) < $\varepsilon$ (\textcolor{green}{true}). The $\alpha$ shows the data and knowledge contributions for the samples.}
    \label{fig:vis}
\end{figure}

Figure \ref{fig:vis} illustrates the final KSAT layer representations (the $Z_{KCLS}$ vectors) and $\alpha_{l}$ values of sample test posts to visualize data and knowledge contexts (see section \ref{subsubsec:layer_output}).

\section{Conclusion}
We proposed KSAT that integrates multiple contexts and data and shows its utility in the domain-specific use-case of MHDA. Although we test KSAT on MHDA, KSAT applies to other domain-specific tasks that require contextualization from multiple knowledge sources. The architecture of KSAT allows for precise measurement and visualization of the contributions from the different knowledge contexts and the data, thus addressing the research aims \textbf{R1} and \textbf{R2}. In future work, we will apply KSAT to other downstream domain-specific knowledge-intensive tasks such as conversational question answering that requires drawing context from multiple knowledge sources. We will replicate KSAT's qualitative evaluation in MHDA for robust human-understandable evaluation on future tasks.

\section{Acknowledgement}

This work is built on prior work \cite{roy2023process,roy2023demo,roy2021depression,roy2021knowledge,roy2021bknowledge,asawa2020covid,venkataramanan2023cook,roy2023knowledge,gaur2021can,roy2023proknow,tsakalidis2022overview,gupta2022learning,dolbir2021nlp,rawte2022tdlr,lokala2021edarktrends,zi2023ierl}, and supported by the National Science Foundation under Grant 2133842, “EAGER: Advancing Neuro-symbolic AI with Deep Knowledge-infused Learning" \cite{sheth2023neurosymbolic,sheth2021knowledge,sheth2022process}.
\bibliographystyle{unsrt}
\bibliography{references}
\appendix
\section{Constructing k-MHDA}\label{subsec:kmhda}
There are $500$ Reddit posts in the MHDA dataset. The knowledge contexts in the CSSRS can be illustrated as a tree (see Figure \ref{fig:cssrs} (\textbf{d})). We can construct a probabilistic decision tree that takes input post $x$ and outputs an outcome $y$ from among the leaves. We can write the tree in algebraic form as shown in \textbf{Equation} \ref{eqn:prob}. 
\begin{equation}
\scriptsize
P(y\mid x,\{\theta_{i}\}) = \sum_{y\in Outcomes}p_{y}\prod_{i=1}^{3}\sum_{x_{sub} \in x}\Bigg( cos\_sim\bigg(x_{sub}^{R},q_{i}^{R}\bigg) \geq \theta_{i}\Bigg) \geq 0.5
\label{eqn:prob}
\end{equation}
 $p_y$ is the ground truth probability for each outcome. Index $i$ iterates through the $3$ concepts in Figure \ref{fig:cssrs} (d). $x_{sub}$ denotes a sub-fragment of the input post ($1$ sentence, $2$ sentence, etc.). $q_i$ denotes the concept texts from the $3$ concepts that $i$ indexes. $x_{sub}^R$ and $q_i^R$ are representations of the post sub-fragment and the concept texts using the sentence-transformer published by Reimers et al. \cite{reimers-2019-sentence-bert}. 
 
 \textbf{Equation} \ref{eqn:sat} determines the presence or absence of concept $q_i$ in a post sub-fragment $x_{sub}$. First, we compute the cosine similarity between their sentence-transformer representations $x_{sub}^R$ and $q_i^R$. If the resulting value is $\geq \theta_i$, we determine that the concept $q_i$ is present in $x_{sub}$, else we determine that the concept $q_i$ is absent in $x_{sub}$.
 
 $\sum_{x_{sub} \in x}(.) \geq 0.5$ in Equation \ref{eqn:prob} is the algebraic form of the $\lor$ operation as we determine that concept $q_i$ is present in the post $x$, if any of the post fragments $x_{sub} \in x$ show presence of concept $q_i$. 
\begin{equation}
\Bigg(cos\_sim\bigg(x_{sub}^{R},q_{i}^{R}\bigg) \geq \theta_{i}\Bigg)
\label{eqn:sat}
\end{equation}
We can then evaluate the Bernoulli Loss $\mathcal{L}$ given an input, outcome pair $(x,y)$ and parameters $\{\theta_{i}\}$  as:
\begin{equation}
\begin{aligned}
\mathcal{L}(x,y,\{\theta_{i}\}) = P(y\mid x,\{\theta_{i}\})log(P(y\mid x,\{\theta_{i}\})) 
+ \\
(1-P(y\mid x,\{\theta_{i}\}))log(1-P(y\mid x,\{\theta_{i}\}))
\end{aligned}
\label{eqn:loss}
\end{equation}
We use grid-search to find a configuration of parameters $\{\theta_i\}$ and post sub-fragment $x_{sub}$ that has the maximum value for

$\prod_{(x,y) \in \mathbf{MHDA}}\mathcal{L}(x,y,\{\theta_{i}\})$. We vary each individual $\theta_i$ in the range $-1$ to $1$ (the \textit{range of the cosine function}) and $x_{sub}$ takes values from the set $\{1,2,3\}$.

Inference is carried out as it is in a decision tree classifier with the concept presence or absence at each branch, evaluated using \textbf{Equation} \ref{eqn:sat}.

\paragraph{\textbf{Knowledge Context Annotation with outputs from grid-search:}} The grid-search yielded outputs $\{\theta_i\} = \{0.3,0.5,0.3\}$, and post sub-fragment size $|x_{sub}| = 1$ (one sentence). Therefore the post ``I don't feel like waking up and have a gun. Oh well.'' is annotated with the knowledge context: \textbf{(Concept 1 (T)), Concept 2 (T), Concept 3 (T)} = \textit{Behavior or Attempt}, as evaluation of \textbf{Equation} \ref{eqn:sat} determines absence of \textbf{Concept 1}, \textbf{Concept 2}, and \textbf{Concept 3} in the post sentence``I don't feel like waking up and have a gun''. The evaluation uses the grid-search outputs of $\{\theta_i\}$. The second sentence ``Oh well' is not necessary to evaluate as we determine a concept's presence or absence in the post if any of the post fragments $x_{sub}$ (one sentence) show the presence of the concept. 
\section{Construction of the \textbf{K-type(i)} and \textbf{K-type(ii)} baseline models}\label{sec:baselines}
For our baseline implementations we adapt the state-of-the-art models for each of the model types \textbf{K-type(i)} and \textbf{K-type(ii)} in Section \ref{subsec:rel_work} for our task description (see Section \ref{subsec:resources}. Specifically, we use the model KnowBERT and K-Adapter as \textbf{K-type(i)} and \textbf{K-type(ii)} model instances respectively \cite{peters2019knowledge,wang2020k}. For knowledge graph embeddings in the KnowBERT case, we utilize TransE embeddings. Similar to their work, we use the RoBERTa model for the task-specific adapter module. We retain all hyperparameters from the original implementations \cite{bordes2013translating,liu2019roberta}.
\end{document}